\documentclass{article}

\usepackage{spconf,amsmath}
\usepackage{hhline}
\usepackage{scalerel} 

\usepackage[utf8]{inputenc} 
\usepackage[T1]{fontenc}    
\usepackage{hyperref}       
\usepackage{url}            
\usepackage{booktabs}       
\usepackage{amsfonts}       
\usepackage{nicefrac}       
\usepackage{microtype}      

\usepackage{subcaption}
\usepackage{multicol}
\usepackage{algorithm}
\usepackage{epsfig}
\usepackage{graphicx}
\usepackage{amsmath}
\usepackage{amssymb}
\usepackage{booktabs} 

\title{Learned Multi-Resolution Variable-Rate Image Compression\\with Octave-based Residual Blocks}

\name
{Mohammad Akbari$^*$, Jie Liang$^*$, Jingning Han$^\dagger$, Chengjie Tu$^\ddagger$}
\address{akbari@sfu.ca, jiel@sfu.ca, jingning@google.com, chengjietu@tencent.com\\Simon Fraser University, Canada$^*$, Google Inc.$^\dagger$, Tencent Technologies$^\ddagger$}

\begin{document}\sloppy

\def\x{{\mathbf x}}
\def\L{{\cal L}}

%

\maketitle

\begin{abstract}
Recently deep learning-based image compression has shown the potential to outperform traditional codecs. However, most existing methods train multiple networks for multiple bit rates, which increase the implementation complexity. In this paper, we propose a new variable-rate image compression framework, which employs generalized octave convolutions (GoConv) and generalized octave transposed-convolutions (GoTConv) with built-in generalized divisive normalization (GDN) and inverse GDN (IGDN) layers. Novel GoConv- and GoTConv-based residual blocks are also developed in the encoder and decoder networks. Our scheme also uses a stochastic rounding-based scalar quantization. To further improve the performance, we encode the residual between the input and the reconstructed image from the decoder network as an enhancement layer. To enable a single model to operate with different bit rates and to learn multi-rate image features, a new objective function is introduced. Experimental results show that the proposed framework trained with variable-rate objective function outperforms the standard codecs such as H.265/HEVC-based BPG and state-of-the-art learning-based variable-rate methods.
\end{abstract}

\section{Introduction}
\label{Introduction}

In the last few years, deep learning has made tremendous progresses in the well-studied topic of image compression. Deep learning-based image compression \cite{akbari2019dsslic,johnston2017improved,li2019learning,minnen2018joint,theis2017lossy,toderici2017full, li2020deep} has shown the potential to outperform standard codecs such as JPEG2000, the H.265/HEVC-based BPG image codec \cite{bellard2017bpg}, and also the new versatile video coding test model (VTM) \cite{vvc-vtm,akbari2020generalized}, making it a very promising tool for the next-generation image compression.

In traditional compression methods, many components are fixed and hand-crafted such as linear transform and entropy coding. Deep learning-based approaches have the potential of automatically exploiting the data features; thereby achieving better compression performance. In addition, deep learning allows non-linear transform coding and more flexible context modellings \cite{balle2020nonlinear}. Various learning-based image compression frameworks have been proposed in the last few years. 

In \cite{balle2016,balle2016end}, a scheme involving a generalized divisive normalization (GDN)-based nonlinear analysis transform, a uniform quantizer, and an inverse GDN (IGDN)-based synthesis transform was proposed. The encoding network consists of three stages of convolution and GDN layers. The decoding network consists of three stages of transposed-convolution and IGDN layers. Despite its simple architecture, it outperforms JPEG2000 in both PSNR and SSIM.

A compressive auto-encoder framework with residual connection as in ResNet (residual neural network) was proposed in \cite{theis2017lossy}, where the quantization was replaced by smooth approximation, and a scaling approach was used to get different rates. In \cite{agustsson2017soft}, a soft-to-hard vector quantization approach was introduced, and a unified framework was developed for both image compression and neural network model compression. 

\begin{figure*}
 \centering
\begin{subfigure}[b]{0.85\linewidth}
 \centering
  \centerline{\includegraphics[width=\textwidth]{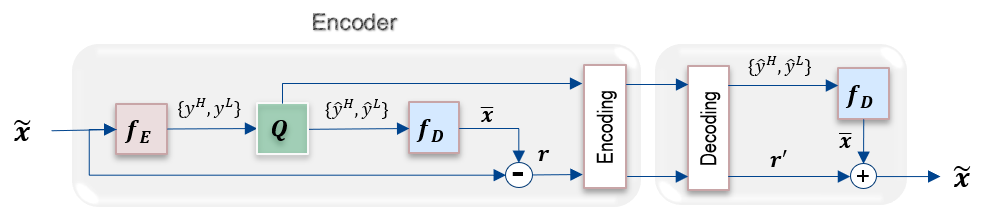}}
\end{subfigure}
\caption{Overall framework of the proposed codec. \textbf{$x$}: input image; \textbf{$f_E$}: deep encoder; \textbf{$\{{y}^H,{y}^L\}$}: factorized code map (including high and low resolution maps); \textbf{$Q$}: uniform scalar quantizer; \textbf{$\{\hat{y}^H,\hat{y}^L\}$}: quantized code map; \textbf{$f_D$}: deep decoder; \textbf{$\bar{x}$}: generated image by the deep decoder; \textbf{$r$}: residual image; 
 \textbf{$r'$}: decoded residual image; \textbf{$\Tilde{x}$}: final reconstructed image.}
\label{fig:framework}
\end{figure*} 

In \cite{akbari2019dsslic}, a deep semantic segmentation-based layered image compression (DSSLIC) was proposed, by taking advantage of the Generative Adversarial Network (GAN). A low-dimensional representation and segmentation map of the input along with the residual between the input and the synthesized image were encoded. It outperforms the BPG codec (RGB4:4:4) in both PSNR and MS-SSIM \cite{wang2003multiscale}.

Most previous works used fixed entropy models shared between the encoder and decoder. In \cite{balle2018variational}, a conditional entropy model based on Gaussian scale mixture (GSM) was proposed where the scale parameters were conditioned on a hyper-prior learned using a hyper auto-encoder. The compressed hyper-prior was transmitted and added to the bit stream as side information. The model in \cite{balle2018variational} was extended in \cite{minnen2018joint,lee2018context} where a Gaussian mixture model (GMM) with both mean and scale parameters conditioned on the hyper-prior was utilized. In these methods, the hyper-priors were combined with auto-regressive priors generated using context models, which outperformed BPG in terms of both PSNR and MS-SSIM. These approaches are jointly optimized to effectively capture the spatial dependencies and probabilistic structures of the latents, which lead to a significant compression performance. However, some of these latents are spatially redundant. In \cite{akbari2020generalized}, a learned multi-resolution image compression approach was proposed that uses generalized octave convolutions (GoConv) and generalized octave transposed-convolutions (GoTConv) to factorize the latents into high and low resolutions. As a result, the spatial redundancy corresponding to the latents is reduced, which improves the compression performance. 

Since most learned image compression methods need to train multiple networks for multiple bit rates, variable-rate image compression approaches have also been proposed in which a single neural network model is trained to operate at multiple bit rates. This approach was first introduced in \cite{toderici2015variable}, which was then generalized for full-resolution images using deep learning-based entropy coding in \cite{toderici2017full}.

In \cite{toderici2015variable}, long short-term memory (LSTM)-based recurrent neural networks (RNNs) and residual-based layer coding was used to compress thumbnail images. Better SSIM results than JPEG and WebP were reported. This approach was generalized in \cite{toderici2017full}, which proposed a variable-rate framework for full-resolution images by introducing a gated recurrent unit, residual scaling, and deep learning-based entropy coding. This method can outperform JPEG in terms of PSNR.

In \cite{cai2018efficient}, a CNN-based multi-scale decomposition transform was optimized for all scales. Rate allocation algorithms were also applied to determine the optimal scale of each image block. The results in \cite{cai2018efficient} were reported to be better than BPG in MS-SSIM. In \cite{zhang2018learned}, a learned progressive image compression model was proposed, in which bit-plane decomposition was adopted. Bidirectional assembling gated units were also introduced to reduce the correlation between different bit-planes \cite{zhang2018learned}.

In \cite{akbari2020learned}, we proposed a variable-rate image compression scheme, by applying GDN-based residual blocks as in ResNet and two novel multi-bit objective functions. The residual between the input and the reconstructed image was also encoded by BPG to further improve the performance. Experimental results show that our variable-rate model can outperform state-of-the-art learning-based variable-rate methods.

In this paper, we extend and improve our previous approach in \cite{akbari2020learned} and propose a new deep learning-based multi-resolution variable-rate image compression framework, which employs GoConv and GoTConv layers developed in \cite{akbari2020generalized} to factorize all the feature maps into high resolution (HR) and low resolution (LR) information. Two novel types of octave-based residual sub-networks are also developed in the encoder and decoder networks, by incorporating separate shortcut connection for HR and LR feature maps. Our scheme uses the stochastic rounding-based scalar quantization as in \cite{toderici2015variable,gupta2015deep,Raiko15}. As in \cite{akbari2020learned}, to further improve the performance, we encode the residual between the input and the reconstructed image from the decoder network by BPG as an enhancement layer. To enable a single model to operate with different bit rates and to learn multi-rate image features, a new variable-rate objective function is introduced \cite{akbari2020learned}. 
Experimental results show that the proposed framework trained with variable-rate objective function outperforms the standard codecs including H.265/HEVC-based BPG in the more challenging YUV4:2:0 and YUV4:4:4 formats, as well as state-of-the-art learning-based variable-rate methods in terms of MS-SSIM metric.

This paper is organized as follows. The architecture of the proposed deep encoder and decoder networks will be described in Section \ref{Variable Chapter: Network Architecture}. Following that, the formulation of the deep encoder and decoder are presented in Sections \ref{Variable Chapter: Deep Encoder} and \ref{Variable Chapter: Deep Decoder}. In Section \ref{Variable Chapter: Objective}, the objective functions are formulated and explained. Finally, we will present the experimental results on Kodak image set as well as the ablation studies in Section \ref{Variable Chapter: Results}.

\begin{figure*}
 \centering
\begin{subfigure}[b]{0.95\linewidth}
 \centering
  \centerline{\includegraphics[width=\textwidth]{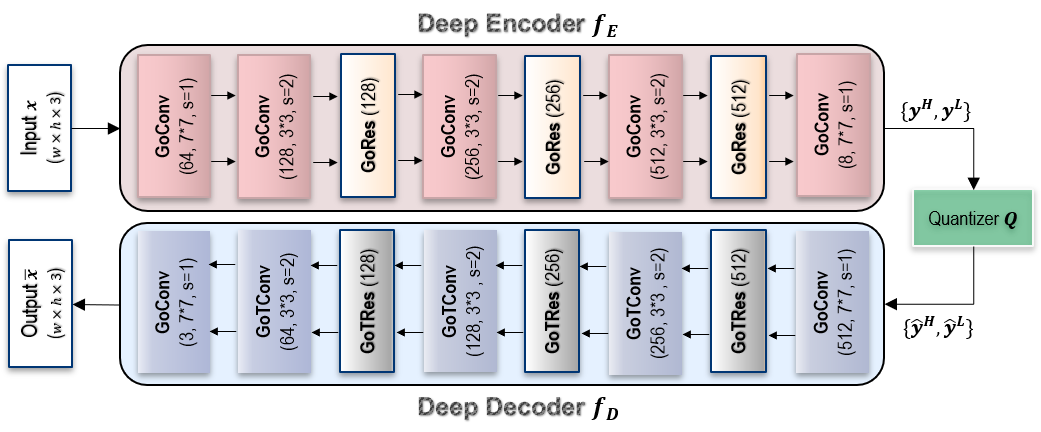}}
\end{subfigure}
\caption{Architecture of the proposed deep encoder and deep decoder networks. \textbf{GoConv/GoTConv (n, k$\times$k, s)}: generalized octave convolutions and transposed-convolutions with n filters of size k$\times$k and stride of s. \textbf{GoRes/GoTRes (n)}: GoConv- and GoTConv-based residual blocks with n filters.
}
\label{fig:deepframework}
\end{figure*}

\section{The Proposed Method}
\label{Proposed Model}

The overall framework of the proposed codec is shown in Fig. \ref{fig:framework}. At the encoder side, two layers of information are encoded: the encoder network output (code map) and the residual image. The code map is composed of HR and LR parts denoted by $\{y^H,y^L\}$, which are obtained by the deep encoder $f_E$. The maps are quantized by a uniform scalar quantizer $Q$, and then separately encoded by the FLIF lossless codec \cite{sneyers2016flif}. The reconstruction of the input image (denoted by $\bar{x}$) is obtained from the quantized maps $\{\hat{y}^H,\hat{y}^L\}$ by the deep decoder $f_D$. To further improve the performance, the residual $r$ between the input and the reconstruction is encoded by the BPG codec as an enhancement layer \cite{akbari2019dsslic}. At the decoder side, the reconstruction $\bar{x}$ from the deep decoder and the decoded residual image $r'$ are added to get the final reconstruction $\tilde{x}$.

\subsection{Network Architecture}
\label{Variable Chapter: Network Architecture}

It has been shown that end-to-end optimization of a model including cascades of differentiable nonlinear transforms has better performance over the traditional linear transforms \cite{balle2016}. One example is the GDN, which is very efficient in gaussianizing local statistics of natural images and has been shown to improve the efficiency of transforms compared to other popular nonlinearities such as ReLU \cite{balle2018efficient}. GDN also provides significant improvements when utilized as a prior for different computer vision tasks such as image denoising and image compression. GDN transforms were first introduced in \cite{balle2016} for a learning-based image compression framework, which had a simple architecture of some down-sampling convolution, each is followed by a GDN layer. 

In order to take the advantage of multi-resolution image compression as well as GDN operations, we incorporate the GoConv and GoTConv architectures \cite{akbari2020generalized} with built-in GDN and IGDN layers in our framework. By using GoConv and GoTConv, the feature representations are factorized into HR and LR maps, where the LR part is represented by a lower spatial resolution to reduce its spatial redundancy. The architecture of the proposed deep encoder and decoder networks are illustrated in Fig. \ref{fig:deepframework}. 
For deeper learning of image statistics and faster convergence, we consider introduce the concept of identity shortcut connection in the ResNet \cite{he2016deep} to some GoConv and GoTConv layers. The architecture of the proposed GoConv- and GoTConv-based residual blocks, which are respectively denoted by GoRes and GoTRes, are shown in Fig. \ref{fig:ResBlocks}. Unlike the traditional residual blocks where Vanilla convolutions followed by ReLU and batch (or instance) normalization are employed, we utilize GoConv and GoTConv layers with build-in GDN and IGDN layers in our residual blocks, which provide better performance and faster convergence rate.

In Fig. \ref{fig:deepframework}, the encoder can be divided into 5 stages. The first and the last GoConv layers are of size 7$\times$7 with stride 1. 
Between them, there are three stages, where each of them includes a 3$\times$3 GoConv layers with stride 2 and a GoRes block. To avoid edge effects, reflection padding of size 3 is used before all convolutions at the first and the last stages. The channel sizes of the convolution layers are 64, 128, 256, 512, and 8, respectively. The deep encoder encodes the input RGB image of size $w\times h\times 3$ into a factorized code map.

The deep decoder decodes the code map back to a reconstructed image. This network is basically the reverse of the deep encoder, where the GoConv and GoRes blocks are respectively replaced by GoTConv and GoTRes. Similar to encoder, reflection padding is used before the convolutions at the first and last stages. The channel sizes in the deep decoder's convolution layers are 512, 256, 128, 64, and 3, respectively.

\subsection{Deep Encoder}
\label{Variable Chapter: Deep Encoder}

\begin{figure*}
\centering
\begin{subfigure}[b]{0.49\linewidth}
 \centering
  \centerline{\includegraphics[width=\textwidth]{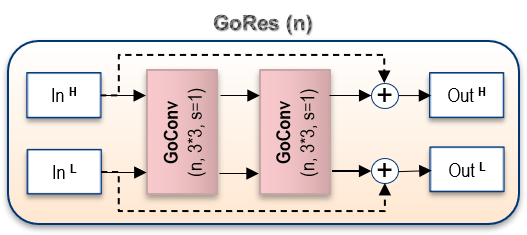}}
\end{subfigure}
\begin{subfigure}[b]{0.49\linewidth}
 \centering
  \centerline{\includegraphics[width=\textwidth]{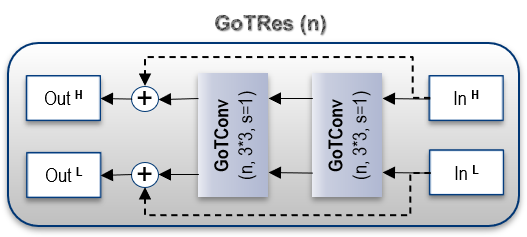}}
\end{subfigure}
\caption{The proposed GoRes and GoTRes transforms. \textbf{n}: the channel size used in the GoConv and GoTConv convolutions;}
\label{fig:ResBlocks}
\end{figure*}

Let $x \in \mathbb{R}^{h\times w\times 3}$ be the original image, the HR and LR code maps $y^H \in \mathbb{R}^{\frac{h}{8} \times \frac{w}{8} \times 8(1-\alpha)}$ and $y^L \in \mathbb{R}^{\frac{h}{4} \times \frac{w}{4} \times 8 \alpha}$ are generated by the parametric deep encoder $f_{E}$ represented as: $\{y^H,y^L\}=f_{E}(x;\Phi)$,
where $\{y^H,y^L\}$ is the code map factorized into HR and LR terms, and $\Phi$ is the parameter vector that needs to be optimized. The encoder consists of 5 GoConv layers. Given the input HR and LR maps $\{I^H,I^L\}$, the output HR and LR feature maps in GoConv are formulated as follows \cite{akbari2020generalized}:
\begin{equation}
\begin{split}
       O^{H} = O^{H\rightarrow H}+g_{\uparrow2}(O^{L\rightarrow L};\Phi^{L\rightarrow H}), \\
       O^{L} = O^{L\rightarrow L}+f_{\downarrow2}(O^{H\rightarrow H};\Phi^{H\rightarrow L}), \\
      \text{ with }  
      O^{H\rightarrow H} =f(I^H;\Phi^{H\rightarrow H}),\\
      O^{L\rightarrow L}=f(I^L;\Phi^{L\rightarrow L}),
\end{split}
\label{eq:conv}
\end{equation}
where $f$ is Vanilla convolution, and $f_{\downarrow2}$ and $g_{\uparrow2}$ are respectively Vanilla convolution and transposed-convolution with stride of 2. $O^{H\rightarrow H}$ and $O^{L\rightarrow L}$ are intra-resolution operations that are used to update the information within each of HR and LR parts. $Y^{H\rightarrow L}$ and $Y^{L\rightarrow H}$ denote inter-resolution communication that enables information exchange between the two parts.
$[\Phi^{H\rightarrow H}, \Phi^{L\rightarrow H}]$ and $[\Phi^{L\rightarrow L}, \Phi^{H\rightarrow L}]$ are the GoConv kernels respectively used for intra- and inter-resolution operations. 

The input to the encoder is not represented as a multi-resolution tensor. So, to compute the output of the first GoConv layer in the encoder, Equation \ref{eq:conv} is modified as follows:
\begin{equation}
       O^{H} = f(x;\Phi^{H\rightarrow H}),~~~O^{L} = f_{\downarrow2}(O^{H};\Phi^{H\rightarrow L}),
\label{equation:GoConv2}
\end{equation}

Except for the first and last GoConv layers, each GoConv is followed by a GoRes transform.
GoRes is composed of two subsequent pairs of GoConv blocks (Equation \ref{eq:conv}) with separate residual connections for the HR and LR maps (Fig. \ref{fig:ResBlocks}).

\begin{figure}
\begin{minipage}[b]{0.44\linewidth}
 \centering
  \centerline{\includegraphics[width=\textwidth]{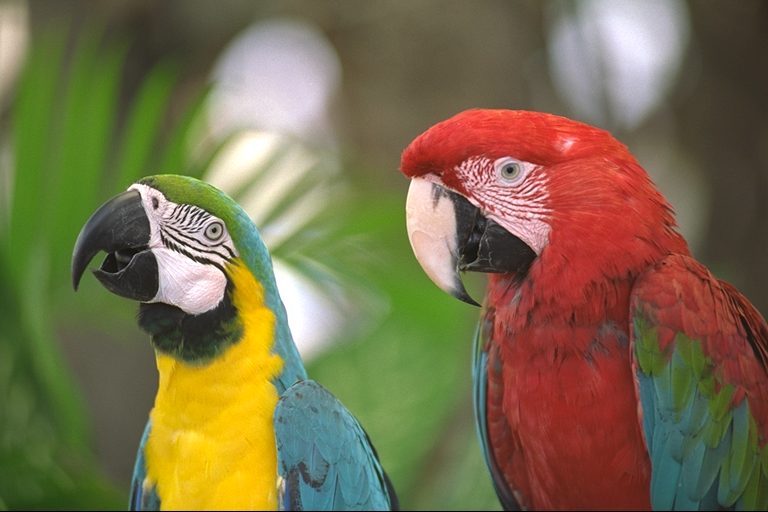}}
\end{minipage}
\begin{minipage}[b]{0.21\linewidth}
 \centering
  \centerline{\includegraphics[width=\textwidth]{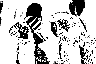}}
  \vspace*{0.05in}
 \centering
  \centerline{\includegraphics[width=\textwidth]{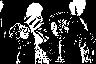}}
\end{minipage}
\begin{minipage}[b]{0.21\linewidth}
 \centering
  \centerline{\includegraphics[width=\textwidth]{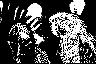}}
  \vspace*{0.05in}
 \centering
  \centerline{\includegraphics[width=\textwidth]{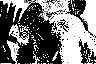}}  
\end{minipage}
\begin{minipage}[b]{0.095\linewidth}
 \centering
  \centerline{\includegraphics[width=\textwidth]{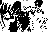}}
  \vspace*{0.05in}
 \centering
  \centerline{\includegraphics[width=\textwidth]{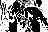}}
  \vspace*{0.05in}
 \centering
  \centerline{\includegraphics[width=\textwidth]{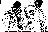}}
  \vspace*{0.05in}
  \centering
  \centerline{\includegraphics[width=\textwidth]{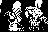}}
\end{minipage}
\caption{Sample quantized code map generated by the deep encoder. \textbf{Left}: original input image from Kodak image set; \textbf{Middle columns}: high resolution feature maps; \textbf{Right column}: low resolution maps.}
\label{fig:code_map}
\end{figure}

\begin{figure*}
\centering
\begin{minipage}[b]{0.49\linewidth}
 \centering
  \centerline{\includegraphics[width=\textwidth]{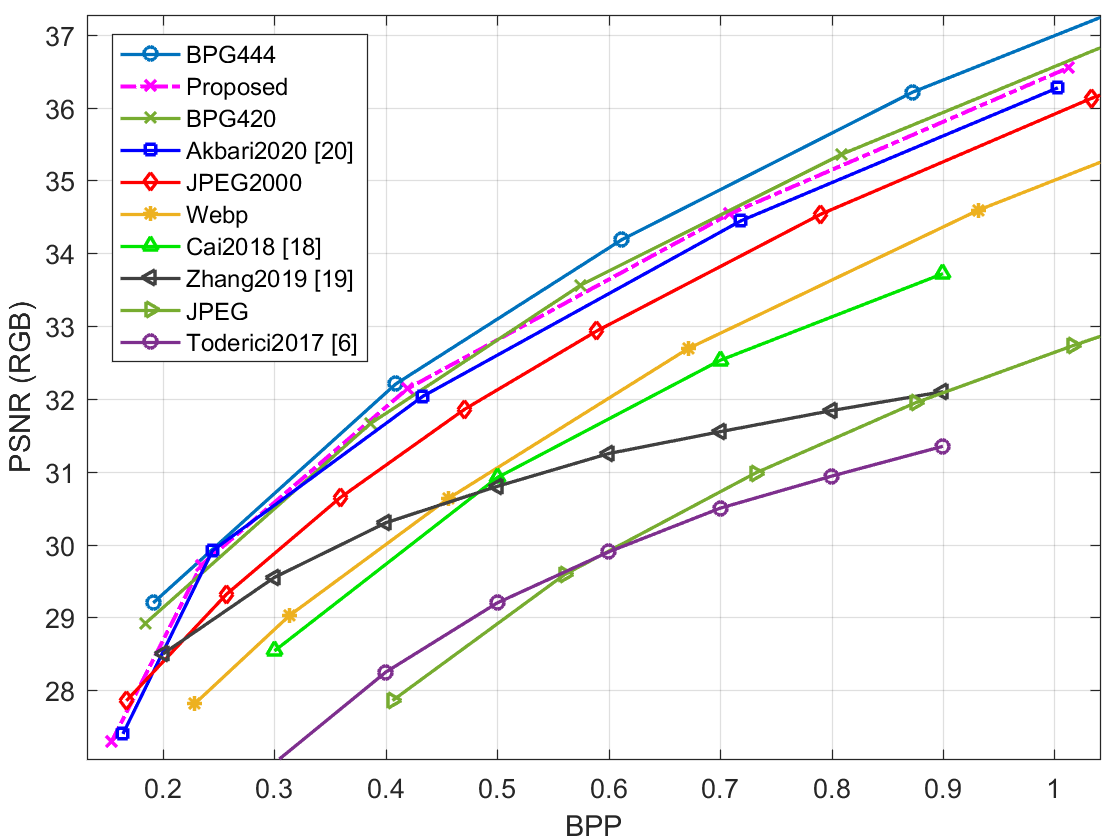}}
\end{minipage}
\begin{minipage}[b]{0.49\linewidth}
 \centering
  \centerline{\includegraphics[width=\textwidth]{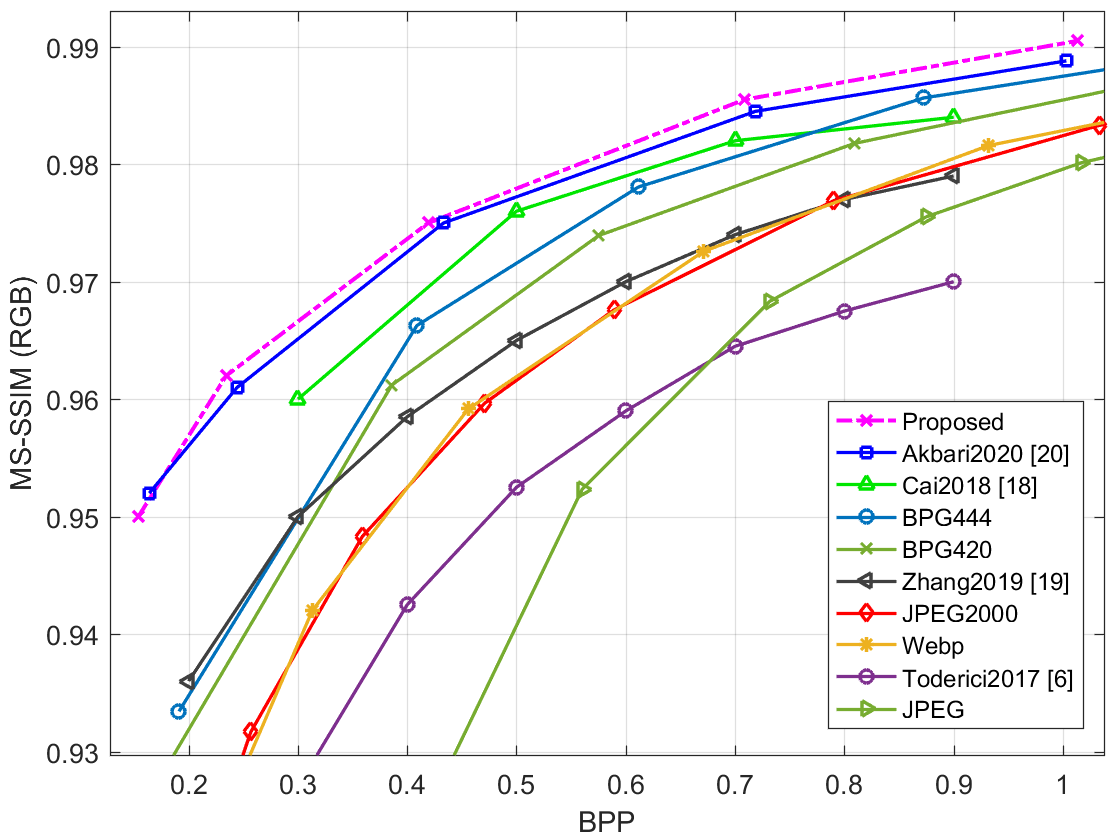}}
\end{minipage}
\caption{Comparison results of proposed variable-rate approach with state-of-the-art variable-rate methods on Kodak test set in terms of PSNR (left) and MS-SSIM (right) vs. bpp (bits/pixel).}
\vskip -3pt
\label{fig:scalable_results_Kodak}
\end{figure*}

\subsection{Stochastic Rounding-Based Quantization}
\label{sec:quantization}

The output of the last stage of the encoder represents the code map $\{y^H,y^L\}$ with $8\alpha$ and $8(1-\alpha)$ channels for HR and LR, respectively. Each HR and LR channel denoted by $\{y_i^{H},y_i^{L}\}$ is then quantized to a discrete-valued vector using a stochastic rounding-based uniform scalar quantizer as: \begin{equation}
\hat{y}_i^{H}=Q(y_i^{H}), \hat{y}_i^{L}=Q(y_i^{L}),
\end{equation}
where the function $Q$ is defined as in \cite{toderici2015variable, gupta2015deep, Raiko15}: 
\begin{equation}
    Q(y_i) = Round\left(\frac{y_i+\epsilon}{\Delta}\right)+z,
\label{eq:quantization}
\end{equation}
where $\epsilon\in[-\frac{1}{2},\frac{1}{2}]$ is produced by a uniform random number generator. $\Delta$ and $z$ respectively represent the quantization step (scale) and the zero-point, which are defined as:
\begin{equation}
    \Delta = \frac{max(y_i)-min(y_i)}{2^B-1},
\label{eq:step}
\end{equation}
and
\begin{equation}
z = \begin{cases}
  0 & \frac{-min(y_i)}{\Delta} < 0,\\
2^B - 1 & \frac{-min(y_i)}{\Delta} > 2^B - 1,\\
  \frac{-min(y_i)}{\Delta} & \text{otherwise,}
\end{cases}
\label{eq:offset}
\end{equation}
where $B$ is the number of bits and $min(y_i)$ and $max(y_i)$ are the input's minimum and maximum values over the $i$th channel, respectively. The zero-point $z$ is an integer ensuring that zero is quantized with no error, which avoids quantization error in common operations such as zero padding \cite{krishnamoorthi2018quantizing}. 

The stochastic rounding approach in Equation \ref{eq:quantization} provides a better performance and convergence rate compared to round-to-nearest algorithm. Stochastic rounding is indeed an unbiased rounding scheme, which maintains a non-zero probability of small parameters \cite{ditheredquantizers}. In other words, it possesses the desirable property that the expected rounding error is zero as follows: 
\begin{equation}
\mathbb{E}\left(Round(x)\right)=x. 
\end{equation}
As a consequence, the gradient information is preserved and the network is able to learn with low bits of precision without any significant loss in performance. 

For the entropy coding of the quantized HR and LR code maps, denoted by $\{\hat{y}^H_{(i)},\hat{y}^L_{(i)}\}$, the FLIF codec \cite{sneyers2016flif} is utilized, which is the state-of-the-art lossless image codec. Since FLIF can also work with grayscale images, each of the quantized code map channels (considered as a grayscale image) is separately entropy-coded by FLIF.

\subsection{Deep Decoder}
\label{Variable Chapter: Deep Decoder}


Given the quantized code maps $\{\hat{y}^{H},\hat{y}^{L}\}$ the parametric decoder $f_{D}$ (with the parameter vector $\Psi$) reconstructs the image $\bar{x} \in \mathbb{R}^{h\times w\times 3}$ as follows: $\bar{x}=f_{D}(\{\hat{y}^H,\hat{y}^L\};\Psi)$.

All the operations performed in the deep encoder are reversed at the decoder side. 
The deep decoder is composed of 5 GoTConv layers defined as follows \cite{akbari2020generalized}: 
\begin{equation}
\begin{split}
       \bar{O}^{H} = \bar{O}^{H\rightarrow H}+g_{\uparrow2}(\bar{O}^{L\rightarrow L};\Psi^{L\rightarrow H}), \\
       \bar{O}^{L} = \bar{O}^{L\rightarrow L}+f_{\downarrow2}(\bar{O}^{H\rightarrow H};\Psi^{H\rightarrow L}) \\
      \text{ with }      \bar{O}^{H\rightarrow H} = g(\bar{I}^H;\Psi^{H\rightarrow H}),\\
      \bar{O}^{L\rightarrow L} = g(\bar{I}^L;\Psi^{L\rightarrow L}).
\end{split}  
\label{eq:deconv}
\end{equation}

Similar to the input of the deep encoder, the output of the last GoTConv at the decoder side is a single tensor representation. For this case, Equation \ref{eq:deconv} is accordingly modified as:
\begin{equation}
\begin{split}
       \bar{O} = \bar{O}^{H\rightarrow H}+g_{\uparrow2}(\bar{O}^{L\rightarrow L};\Psi^{L\rightarrow H}), \\
      \text{ with }      \bar{O}^{H\rightarrow H} = g(\bar{I}^H;\Psi^{H\rightarrow H}),\\
      \bar{O}^{L\rightarrow L} = g(\bar{I}^L;\Psi^{L\rightarrow L}),
\end{split}
\label{equation:GoTConv2}
\end{equation}
where $g$ is Vanilla transposed-convolution.

As the reverse of the encoder, each GoTConv at the is followed by a GoTRes block (except for the last two GoTConv layers).
GoTRes consists of two subsequent pairs of an GoTConv operations (Equation \ref{eq:deconv}). The reconstructed image $\bar{x}$ is finally resulted from the output of the decoder.

\begin{figure}
\centering
\begin{minipage}[b]{0.99\linewidth}
 \centering
  \centerline{\includegraphics[width=\textwidth]{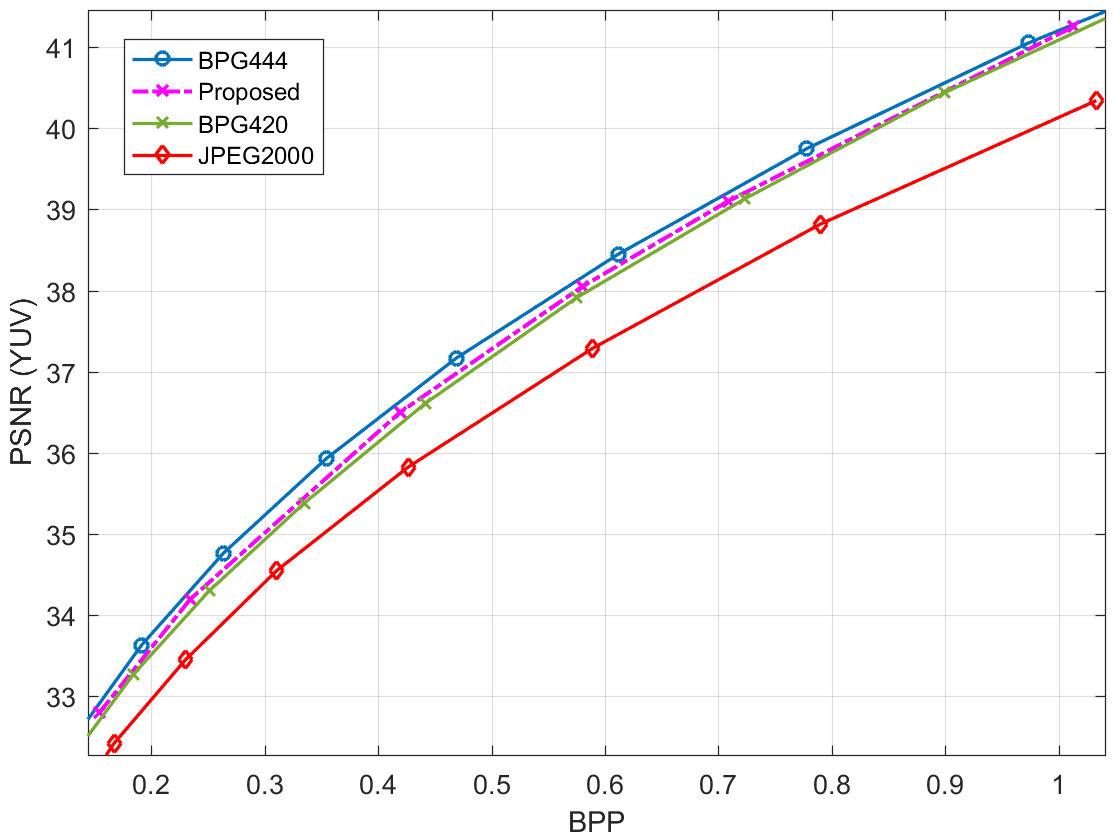}}
\end{minipage}
\caption{Comparison results of proposed variable-rate approach with standard codecs on Kodak test set in terms of PSNR (YUV) vs. bpp (bits/pixel).}
\vskip -3pt
\label{fig:scalable_results_Kodak_yuv}
\end{figure}

\begin{figure*}
\centering
\begin{subfigure}[b]{.32\textwidth}
 \centering
  \centerline{\includegraphics[width=\textwidth]{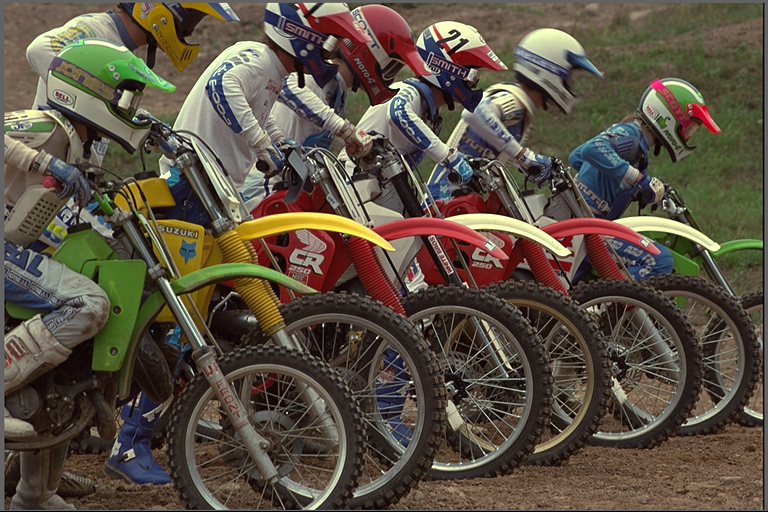}}
  \subcaption{\scriptsize Original}
\end{subfigure}
\begin{subfigure}[b]{.32\textwidth}
 \centering
  \centerline{\includegraphics[width=\textwidth]{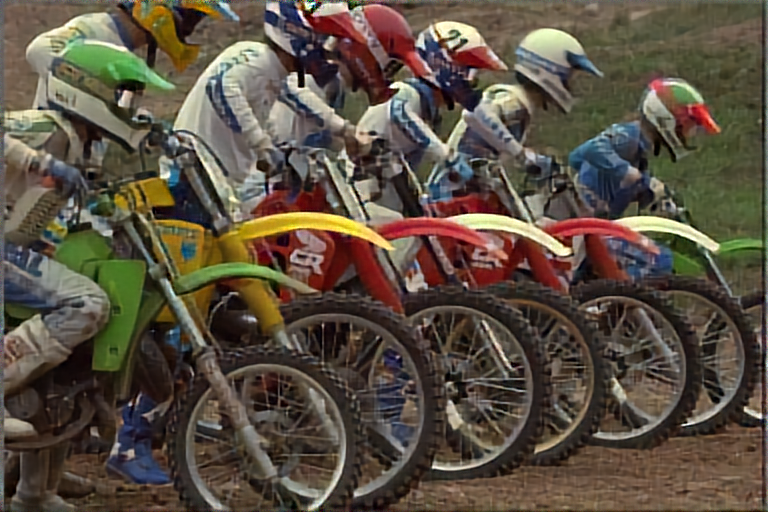}}
  \subcaption{\scriptsize Proposed (0.17bpp, 23.36dB, 0.949)}
\end{subfigure}
\begin{subfigure}[b]{.32\textwidth}
 \centering
  \centerline{\includegraphics[width=\textwidth]{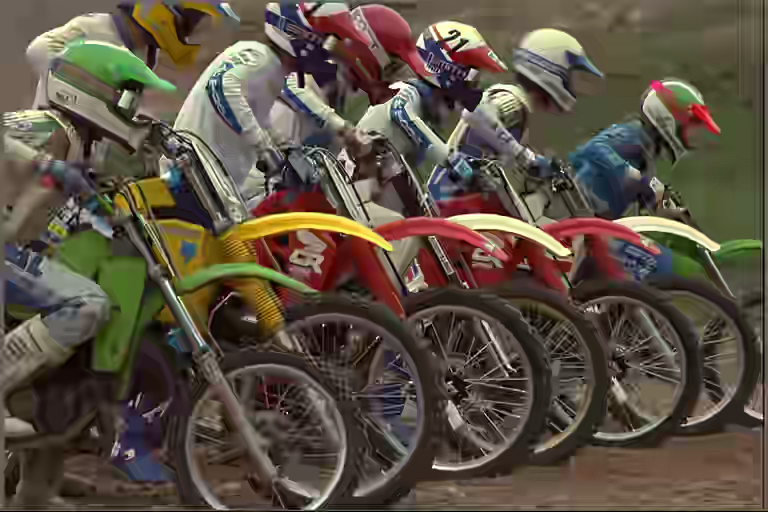}}
  \subcaption{\scriptsize BPG444 (0.18bpp, 23.66dB, 0.891)}
\end{subfigure}
\begin{subfigure}[b]{.32\textwidth}
 \centering
  \centerline{\includegraphics[width=\textwidth]{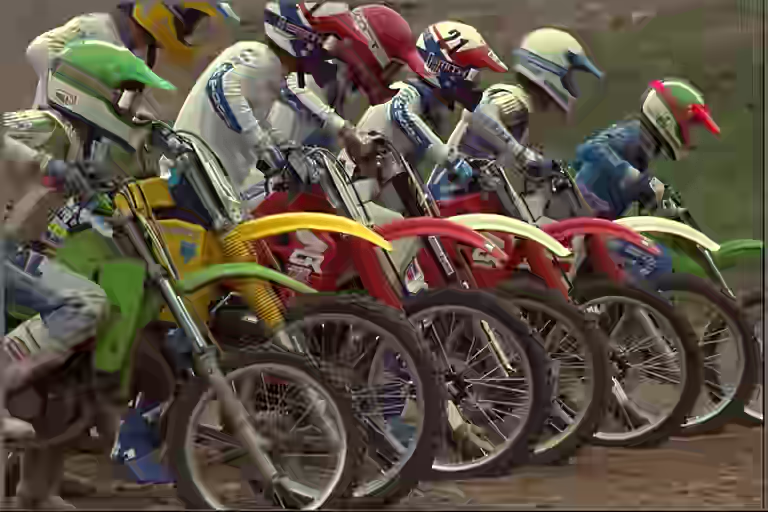}}
  \subcaption{\scriptsize BPG420 (0.18bpp, 23.59dB, 0.888)}
\end{subfigure}
\begin{subfigure}[b]{.32\textwidth}
 \centering
  \centerline{\includegraphics[width=\textwidth]{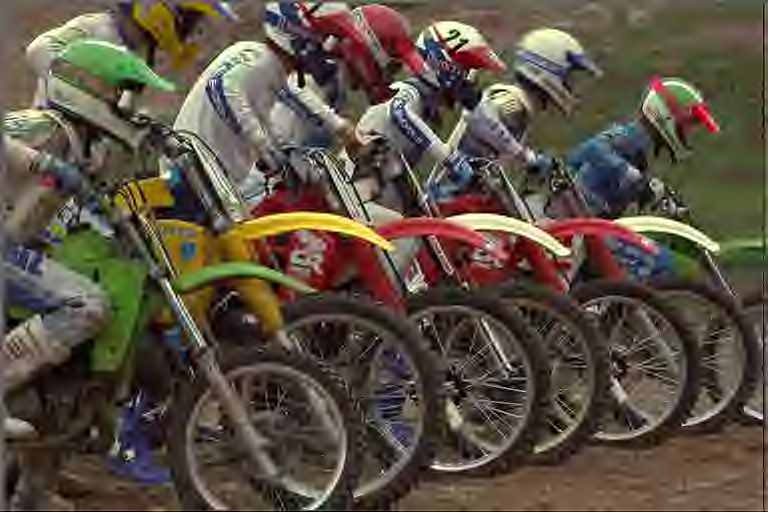}}
  \subcaption{\scriptsize JPEG2000 (0.17bpp, 22.47dB, 0.866)}
\end{subfigure}
\begin{subfigure}[b]{.32\textwidth}
 \centering
  \centerline{\includegraphics[width=\textwidth]{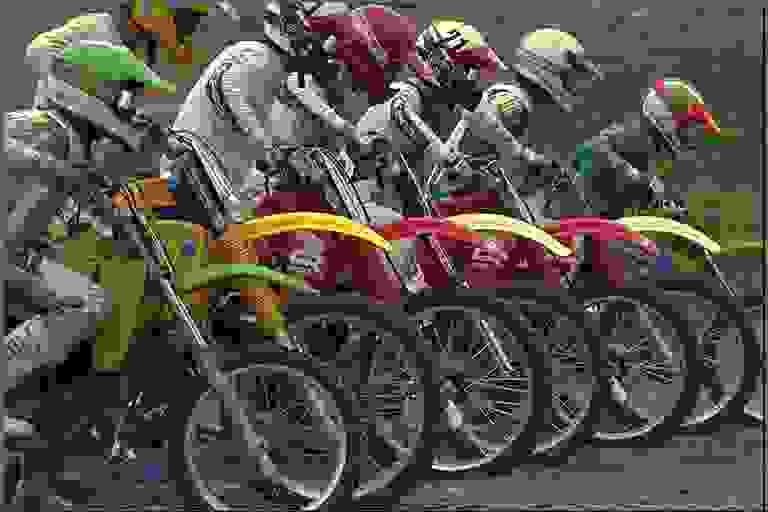}}
  \subcaption{\scriptsize JPEG (0.21bpp, 19.42dB, 0.749)}
\end{subfigure}
\caption{Kodak visual example 1 (bits/pixel, PSNR, MS-SSIM). \textit{BPG444}: YUV (4:4:4) format; \textit{BPG420}: YUV (4:2:0) format.}
\label{fig:variable_kodak_quan1}
\end{figure*}

\begin{figure*}
\centering
\begin{subfigure}[b]{.32\textwidth}
 \centering
  \centerline{\includegraphics[width=\textwidth]{img_kodim23_real_image.png}}
  \subcaption{\scriptsize Original}
\end{subfigure}
\begin{subfigure}[b]{.32\textwidth}
 \centering
  \centerline{\includegraphics[width=\textwidth]{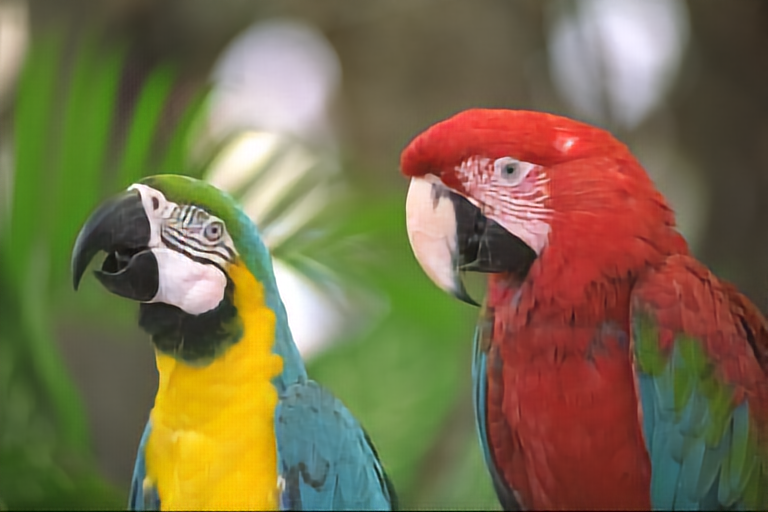}}
  \subcaption{\scriptsize Proposed (0.10bpp, 31.87dB, 0.980)}
\end{subfigure}
\begin{subfigure}[b]{.32\textwidth}
 \centering
  \centerline{\includegraphics[width=\textwidth]{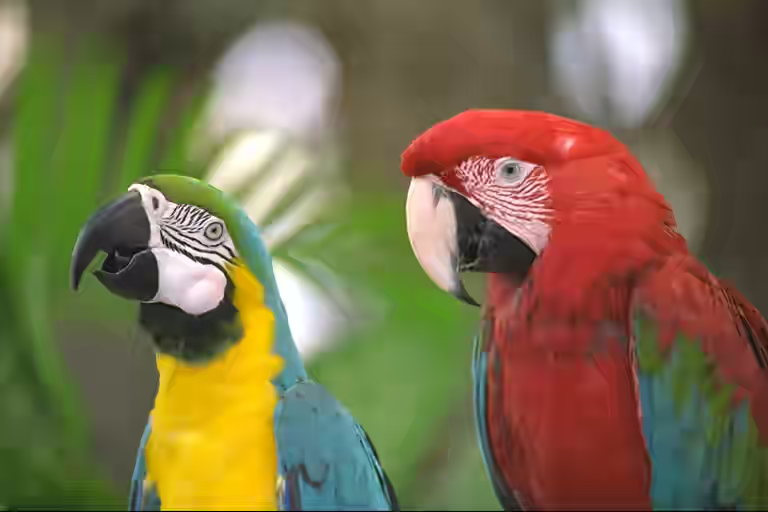}}
  \subcaption{\scriptsize BPG444 (0.10bpp, 32.25dB, 0.949)}
\end{subfigure}
\begin{subfigure}[b]{.32\textwidth}
 \centering
  \centerline{\includegraphics[width=\textwidth]{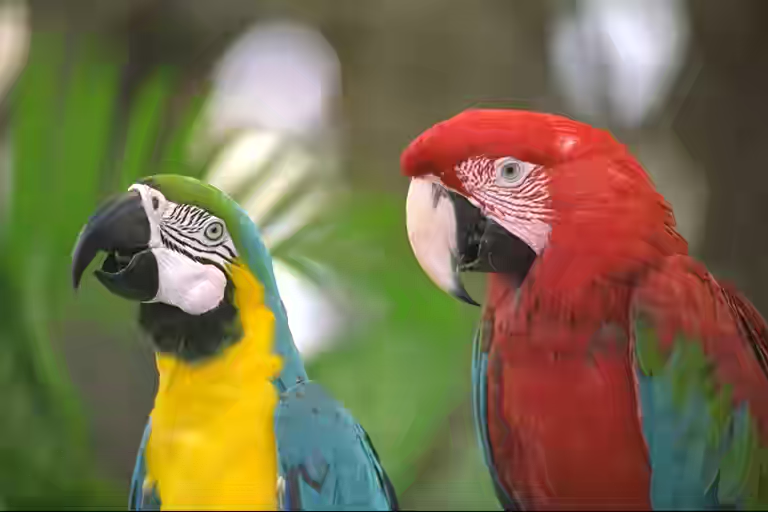}}
  \subcaption{\scriptsize BPG420 (0.11bpp, 32.22dB, 0.948)}
\end{subfigure}
\begin{subfigure}[b]{.32\textwidth}
 \centering
  \centerline{\includegraphics[width=\textwidth]{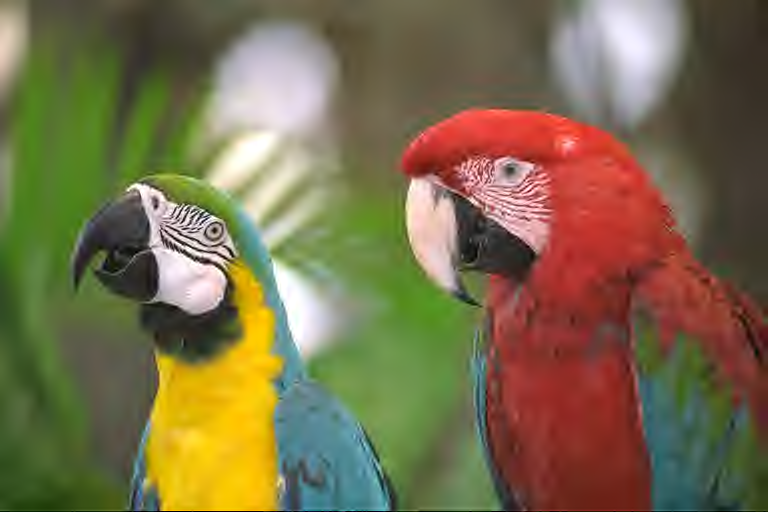}}
  \subcaption{\scriptsize JPEG2000 (0.10bpp, 30.71dB, 0.936)}
\end{subfigure}
\begin{subfigure}[b]{.32\textwidth}
 \centering
  \centerline{\includegraphics[width=\textwidth]{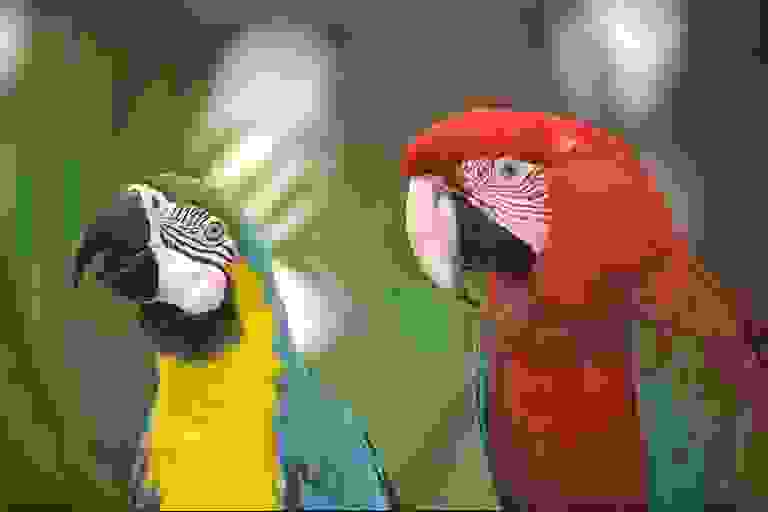}}
  \subcaption{\scriptsize JPEG (0.16bpp, 22.50dB, 0.727)}
\end{subfigure}
\caption{Kodak visual example 2 (bits/pixel, PSNR, MS-SSIM). \textit{BPG444}: YUV (4:4:4) format; \textit{BPG420}: YUV (4:2:0) format.}

\label{fig:variable_kodak_quan2}
\end{figure*}

\subsection{Residual Coding}

As an enhancement layer to the bit-stream, the residual $r$ between the input image $x$ and the deep decoder's output $\bar{x}$ is further encoded by the BPG codec \cite{akbari2019dsslic,akbari2020learned}. To do this, the minimal and the maximal values of the residual image $r$ are first obtained, and the range between them is rescaled to [0,255], so that we can call the BPG codec directly to encode it as a regular 8-bit image. The minimal and maximal values are also sent to decoder for inverse scaling after BPG decoding.

\subsection{Objective Function and Optimization}
\label{Variable Chapter: Objective}

Our cost function is a combination of L2-norm loss denoted by $\mathcal{L}_{2}$, and MS-SSIM loss \cite{wang2004image} denoted by $\mathcal{L}_{MS}$ as follows:
\begin{equation}
\mathcal{L}(\Phi,\Psi) =  2\mathcal{L}_{2}+ \mathcal{L}_{MS},
\end{equation}
where $\Phi$ and $\Psi$ are the optimization parameter vectors of the deep encoder and decoder, respectively, each is defined as a full set of their parameters across all their layers as: $\Phi=\{\Phi^{L\rightarrow H},\Phi^{H\rightarrow L}\}$ and $\Psi=\{\Psi^{L\rightarrow H},\Psi^{H\rightarrow L}\}$.
 In order to optimize the parameters such that our codec can operate at a variety of bit rates, we propose the following novel variable-rate objective functions for the $\mathcal{L}_{2}$ and $\mathcal{L}_{MS}$ losses:
\begin{equation}
\label{eq_obj}
\mathcal{L}_{2}  = \sum_{B\in {R}} {\lVert x-\bar{x}_B\rVert}_2,
\end{equation}
and
\begin{equation}
 \mathcal{L}_{MS} = -\sum_{B\in {R}} I_{M}(x,\bar{x}_B)\prod_{j=1}^M{ C_{j}(x,\bar{x}_B).S_{j}(x,\bar{x}_B)},
\end{equation}
where $\bar{x}_B$ denotes the reconstructed image with $B$-bit quantizer (Equations \ref{eq:step} and \ref{eq:offset}), and $B$ can take all possible values in a set ${R}$. In this paper, ${R}=\{2,4,8\}$ is used for training variable-rate network model. 
The MS-SSIM metric use luminance $I$, contrast $C$, and structure $S$ to compare the pixels and their neighborhoods in $x$ and $\bar{x}$ defined as:
\begin{equation}
\begin{split}
I(x,\bar{x})=\frac{2\mu_x\mu_{\bar{x}}+C_1}{\mu^2_x+\mu^2_{\bar{x}}+C_1},\\
C(x,\bar{x})=\frac{2\sigma_x\sigma_{\bar{x}}+C_2}{\sigma^2_x+\sigma^2_{\bar{x}}+C_2},\\
S(x,\bar{x})=\frac{\sigma_{x\bar{x}}+C_3}{\sigma_x\sigma_{\bar{x}}+C_3},
\end{split}
\end{equation} 
where $\mu_x$ and $\mu_{\bar{x}}$ are the means of $x$ and $\bar{x}$, $\sigma_x$ and $\sigma_{\bar{x}}$ are the standard deviations, and $\sigma_{x\bar{x}}$ is the correlation coefficient. $C1$, $C2$, and $C3$ are the constants used for numerical stability. Moreover, MS-SSIM operates at multiple scales where the images are iteratively downsampled by factors of  $2^{j}$, for $j \in[1,M]$.

Our goal is to minimize the objective $\mathcal{L}(\Phi,\Psi)$ over the continuous parameters $\{\Phi, \Psi\}$. However, both terms depend on the quantized values of $\hat{y}$ whose derivative is discontinuous, which makes the quantizer non-differentiable \cite{balle2016}. To overcome this issue, the fact that the exact derivatives of discrete variables are zero almost everywhere is considered, and the straight-through estimate (STE) approach in \cite{bengio2013estimating} is employed to approximate the differentiation through discrete variables in the backward pass. Using STE, we basically set the incoming gradients to our quantizer equal to its outgoing gradients, which indeed disregards the gradients of the quantizer. The concept of a straight through estimator is that you set the incoming gradients to a threshold function equal to it's outgoing gradients, disregarding the derivative of the threshold function itself.

Note that many methods optimize for PSNR and MS-SSIM separately in order to get better performance in each of them, while our scheme jointly optimizes for both of them, which can still achieve satisfactory results in both metrics.

\begin{figure*}
 \centering
\begin{minipage}[b]{0.49\linewidth}
 \centering
  \centerline{\includegraphics[width=\textwidth]{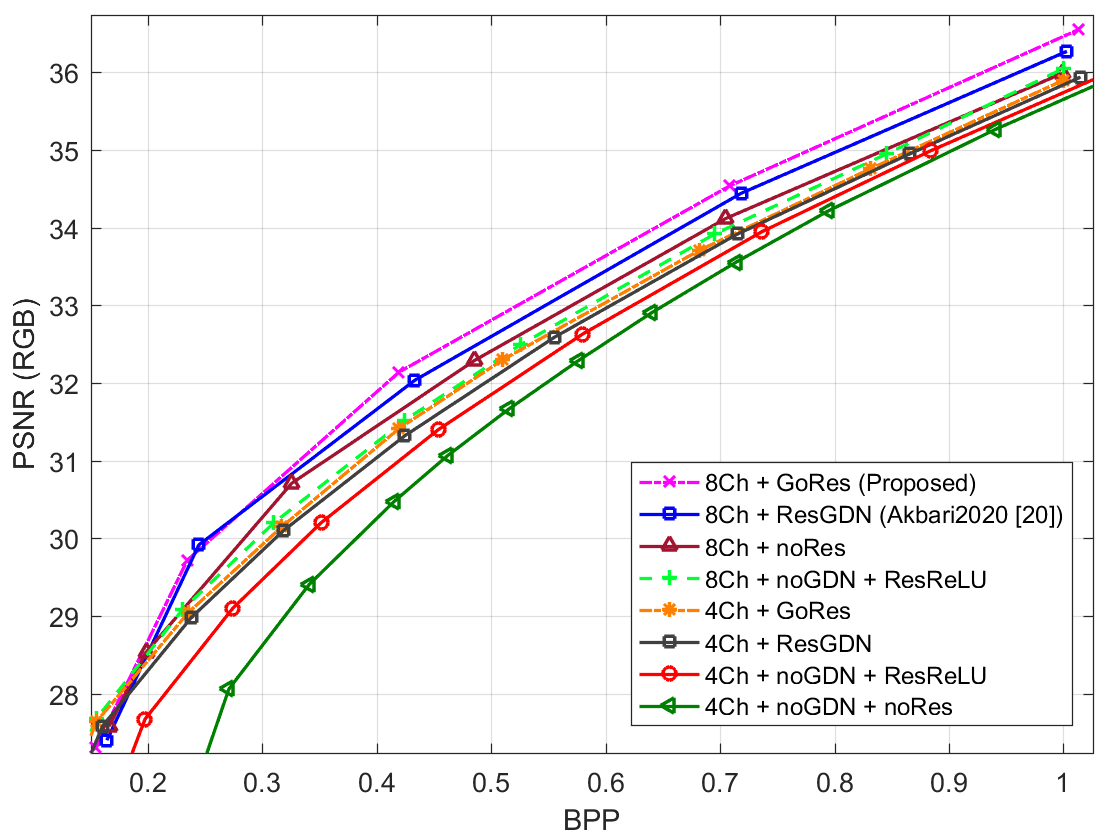}}
\end{minipage}
\begin{minipage}[b]{0.49\linewidth}
 \centering
  \centerline{\includegraphics[width=\textwidth]{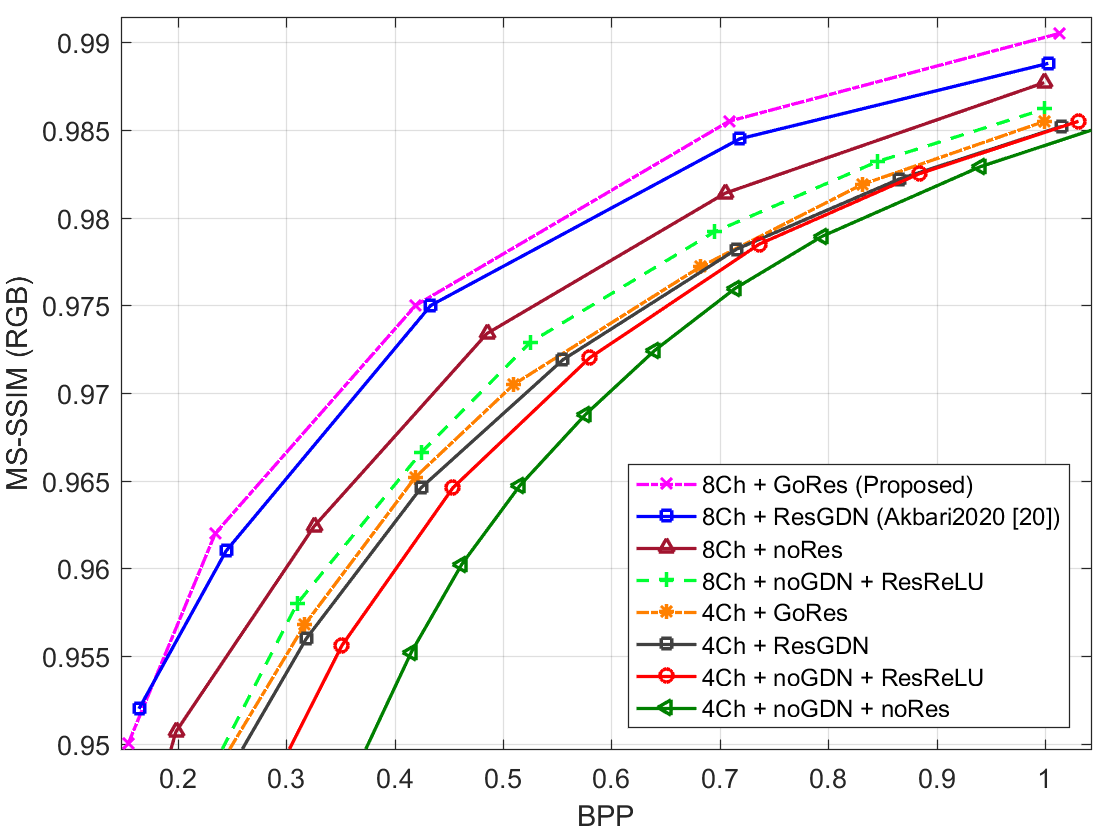}}
\end{minipage}
\caption{Ablation studies with different model configurations. \textbf{\textit{n}Ch}: $n$ channels for the code map; \textbf{ResGDN}: ResGDN/ResIGDN transforms in the network architecture as in \cite{akbari2020learned}; \textbf{ResReLU}: conventional residual block with ReLU; \textbf{noRes}: no residual block is used; \textbf{noGDN}: GDN and IGDN layers in our main architecture replaced by ReLU.}

\label{fig:ablation_results_Kodak}
\end{figure*}

\begin{table}
\centering
\begin{tabular}{c|c|c|}
\cline{2-3}
                                        & \textbf{BD-Rate (\%)} & \textbf{BD-PSNR (dB)} \\ \hline
\multicolumn{1}{|c|}{\textbf{JPEG2000}} & -16.0516              & 0.7686                \\ \hline
\multicolumn{1}{|c|}{\textbf{BPG420}}   & -2.4469               & 0.1132                \\ \hline
\multicolumn{1}{|c|}{\textbf{BPG444}}   & 3.0992                & -0.1345               \\ \hline
\end{tabular}
\caption{Bjontegaard-based comparison results between our method and standard codecs JPEG2000, BPG420, and BPG444. The PSNR results are in YUV.}
\label{tbl:bd}
\end{table}

\section{Experimental Results}
\label{Variable Chapter: Results}

The ADE20K dataset \cite{zhou2017scene} was used for training the proposed model. The images with at least 512 pixels in height or width were used (9272 images in total). All images were rescaled to $h=256$ and $w=256$ to have a fixed size for training. As in \cite{akbari2020generalized}, we set the low resolution ratio $\alpha=0.5$ to respectively get the HR and LR code maps of size 32$\times$32$\times$4 and 16$\times$16$\times$4. The code maps corresponding to one sample image from Kodak image set are shown in Fig. \ref{fig:code_map}. The deep encoder and decoder models were jointly trained for 200 epochs with mini-batch stochastic gradient descent (SGD) and a mini-batch size of 16. The Adam solver with learning rate of 0.00002 was fixed for the first 100 epochs, and was gradually decreased to zero for the next 100 epochs. All the networks were trained in the RGB domain. 
In this section, we compare the performance of the proposed scheme with two types of methods: 1) standard codecs including JPEG, JPEG2000 \cite{christopoulos2000jpeg2000}, WebP \cite{webp2018}, and the H.265/HEVC intra coding-based BPG codec \cite{bellard2017bpg}; and 2) the state-of-the-art learning-based variable-rate image compression methods in \cite{toderici2017full}, \cite{cai2018efficient}, \cite{zhang2018learned}, and \cite{akbari2020learned}, in which a single network was trained to generate multiple bit rates. 
We use both PSNR and MS-SSIM \cite{wang2003multiscale} as the evaluation metrics.

The model was trained using 3 different bit rates, i.e., ${R}=\{2,4,8\}$ in Eq. \ref{eq_obj}. However, the trained model can operate at any bit rate in range $[1,8]$ at the test time. 

The comparison results on the Kodak set (averaged over 24 images) are shown in Fig. \ref{fig:scalable_results_Kodak}. Different points on the R-D curve of our variable-rate results are obtained from 5 different bit rates for the code maps in the base layer, i.e., $R = \{3,4,5,6,7\}$. The corresponding residual images $r$ in the enhancement layer are coded by BPG (YUV4:4:4) with quantizer parameters of $\{50, 40, 35, 30, 25\}$, respectively. Better results can be obtained by performing some rate allocation optimizations.

As shown in Fig. \ref{fig:scalable_results_Kodak}, our method outperforms the state-of-the-art learning-based variable-rate image compression models and JPEG2000 in terms of both PSNR (RGB) and MS-SSIM (RGB). Our PSNR results are slightly lower than BPG (YUV4:4:4) and almost the same as BPG (YUV4:2:0), but we achieve better MS-SSIM, especially at low rates. The comparison results in PSNR (YUV) are also presented in Fig. \ref{fig:scalable_results_Kodak_yuv} in which our method is slightly better than BPG (YUV4:2:0). Table \ref{tbl:bd} summarizes the Bjontegaard (BD)-based average gain in PSNR (YUV) and also average saving bitrates compared to other standard codecs. Our approach has a bitrate saving and PSNR gain of $\approx$16\% and $\approx$0.77dB compared to JPEG2000, and also a saving and gain of $\approx$2.5\% and $\approx$0.12dB compared to BPG (YUV4:2:0).

The BPG-based residual coding in our scheme is exploited to avoid re-training the entire model for another bit rate and more importantly to boost the quality at high bit rates. For the 5 points (low to high) in Fig. \ref{fig:scalable_results_Kodak}, the percentage of bits used by residual image is \{2\%, 34\%, 52\%, 68\%, 76\%\}. This shows that as the bit rate increases, the residual coding has more significant contribution to the R-D performance.




Two visual examples from the Kodak image set are given in Figures \ref{fig:variable_kodak_quan1} and \ref{fig:variable_kodak_quan2} in which our results 
are compared with BPG (YUV4:4:4), BPG (YUV4:2:0), JPEG2000, and JPEG. JPEG has very poor performance due to the ringing artifacts at edges. The BPG has the highest PSNR and also smoother results compared to JPEG2000. However, the details and fine structures in BPG results (e.g., the grooves on the door in Fig. \ref{fig:variable_kodak_quan1} and the grass on the ground in Fig. \ref{fig:variable_kodak_quan2}) are not well-preserved in many areas. Our method achieves the best MS-SSIM and also provides the highest visual quality compared to the other methods including BPG. 

\subsection{Other Ablation Studies}

In order to evaluate the performance of different components of the proposed framework, the ablation studies reported in our previous work \cite{akbari2020learned} are re-discussed and compared with the proposed method. The results are shown in Fig. \ref{fig:ablation_results_Kodak}.
\begin{itemize}
    \item 
\textbf{Code map channel size}: Fig. \ref{fig:ablation_results_Kodak} shows the results with channel sizes of 4 and 8 (i.e., $c\in\{4,8\}$). It can be seen that $c=8$ has better results than $c=4$ in both GoRes and ResGDN cases. In general, we find that a larger code map channel size with smaller quantization bits provide a better R-D performance because deeper texture information of the input image is preserved within the feature maps. 
\item
\textbf{GDN vs. ReLU}: In order to show the performance of the GDN/IGDN transforms, we make a comparison with a ReLU-based variant of our model in \cite{akbari2020learned}, denoted as $noGDN$ in Fig. \ref{fig:ablation_results_Kodak}. In this model, all GDN and IGDN layers in the deep encoder and decoder are removed; instead, instance normalization followed by ReLU are added to the end of all convolution layers. The last GDN and IGDN layers in the encoder and decoder are replaced by a Tahn layer. As shown in Fig. \ref{fig:ablation_results_Kodak}, the models with GDN structure outperform the ones without GDN.
\item
\textbf{Conventional vs. GDN/IGDN-based residual transforms}: In this scenario, the model composed of the ResGDN/ResIGDN transforms proposed in \cite{akbari2020learned} (denoted by ResGDN in Fig. \ref{fig:ablation_results_Kodak}) is compared with the conventional ReLU-based residual block (denoted by $ResReLU$) in which all the GDN/IGDN layers are replaced by ReLU. The results with no residual blocks, denoted by $noRes$, are also included. As demonstrated in Fig. \ref{fig:ablation_results_Kodak}, the models with GoRes or ResGDN achieve better performance compared to the other scenarios without residual blocks.
\end{itemize}


In terms of complexity, the average processing time of the deep encoder and deep decoder on Kodak are $\approx$65ms and $\approx$51ms on a TITAN X Pascal GPU, respectively. The encoding and decoding times for Toderici2017 \cite{toderici2017full} are $\approx$1600ms and $\approx$1000ms. The other previous works have not reported the complexity of their methods.



\section{Conclusion}

In this paper, we proposed a new variable-rate image compression framework, by applying GoConv and GoTConv layers and incorporating the octave-based shortcut connections. We also used a stochastic rounding-based scalar quantization. To further improve the performance, the residual between the input and the reconstructed image from the decoder network was encoded by BPG as an enhancement layer. A novel variable-rate objective function was also proposed. 

Experimental results showed that our variable-rate model can outperform all standard codecs including BPG in MS-SSIM as well as state-of-the-art learning-based variable-rate methods in both PSNR and MS-SSIM. Despite the good MS-SSIM performance of our method compared to other standard codecs, it is not still as good as BPG444 nor VVC in terms of PSNR. Further gains can be achieved by optimizing multiple networks for different bit rates independently. Another future topic is the rate allocation optimization between the base layer and the enhancement layer.

\section*{Acknowledgement}
This work is supported by the Natural Sciences and Engineering Research Council (NSERC) of Canada under grant RGPIN-2015-06522.


\bibliography{refs}
\bibliographystyle{IEEEtran}

\end{document}